\documentclass[final]{siamltex}

\usepackage[parfill]{parskip}    
\usepackage{amsmath,amssymb,latexsym,enumerate,mathrsfs}
\usepackage{hyperref}
\usepackage{array}
\usepackage{times}
\usepackage[]{mdframed}
\usepackage{epstopdf}
\usepackage{subfig}
\usepackage{graphicx}
\usepackage{hyperref} 
\usepackage{multicol}
\usepackage{framed}
\usepackage{moreverb}
\usepackage{marvosym}
\usepackage{color,soul}
\definecolor{lightblue}{rgb}{.90,.95,1}
\sethlcolor{yellow}

\DeclareMathOperator*{\argmin}{arg\,min}
\DeclareMathOperator*{\argmax}{arg\,max}

\title{{\large A brief note}\\ ~\\
On learning problem with global  perspectives}

\author{Getachew K. Befekadu\footnote{\scriptsize Getachew K. Befekadu is with the Department of Electrical \& Computer Engineering, College of Engineering, Physics, and Computing, The Catholic University of America, ~ Washington, DC 20064, USA. E-mail:\,{\tt befekadu@cua.edu}}}

\begin{document}
\maketitle

\renewcommand{\thefootnote}{\arabic{footnote}}

\begin{abstract}
This brief note considers the problem of learning with dynamic-optimizing principal-agent setting, in which the agents are allowed to have global perspectives about the learning process, i.e., the ability to view things according to their relative importances or in their true relations based-on some aggregated information shared by the principal. Whereas, the principal, which is exerting an influence on the learning process of the agents in the aggregation, is primarily tasked to solve a high-level optimization problem posed as an empirical-likelihood estimator under conditional moment restrictions model that also accounts information about the agents' predictive performances on out-of-samples as well as a set of private datasets available only to the principal. In particular, we present a coherent mathematical argument which is necessary for characterizing the learning process behind this abstract principal-agent learning framework, although we acknowledge that there are a few conceptual and theoretical issues still need to be addressed. 
\end{abstract}

\section{Introduction} \label{S1}
In this brief note, we considers the problem of learning with dynamic-optimizing principal-agent setting, in which the agents are allowed to have global perspectives about the learning process, i.e., the ability to view things according to their relative importances or in their true relations based-on some aggregated information shared by the principal. Whereas, the principal, which is exerting an influence on the learning process of the agents in the aggregation, is primarily tasked to solve a high-level optimization problem posed as an empirical-likelihood estimator under conditional moment restrictions model that also accounts information about the agents' predictive performances on out-of-samples as well as a set of private datasets available only to the principal (e.g., see \cite{r1}, \cite{r2}, \cite{r3}, \cite{r4} and \cite{r5} for further discussions on empirical likelihood methods with moment restrictions). Here, we provide a coherent mathematical argument which is necessary for characterizing the learning process behind this abstract dynamic-optimizing principal-agent learning framework. Note that, due to the inherent feedbacks behavior among the agents, the proposed learning framework remarkably offers some advantages in terms of stability and consistency, despite that both the principal and the agents do not necessarily need to have any knowledge of the sample distributions or the quality of each others datasets. Finally, it is worth remarking that such a learning framework can provide new insights in the context of collaborative learning problem with global perspectives that exploits the principal-agent setting (e.g., see \cite{r6}, \cite{r7}, \cite{r8} or \cite{r9} for related discussions), although we acknowledge that there are a number of conceptual and theoretical problems, such as small sample properties, still need to be addressed.

The remainder of this brief note is organized as follows. In Section~\ref{S2}, we provide a formal problem statement, in which we present our learning framework with global perspectives that makes use of a dynamic-optimizing principal-agent setting. In Section~\ref{S3}, we present a coherent mathematical argument which is necessary for characterizing the learning process behind this abstract learning framework. This section also presents a generic algorithm that provides a viable way for implementing such a learning problem with global perspectives.

\section{Problem setup} \label{S2} 
\begin{figure}[h]
\begin{center}
 \includegraphics[scale=0.55]{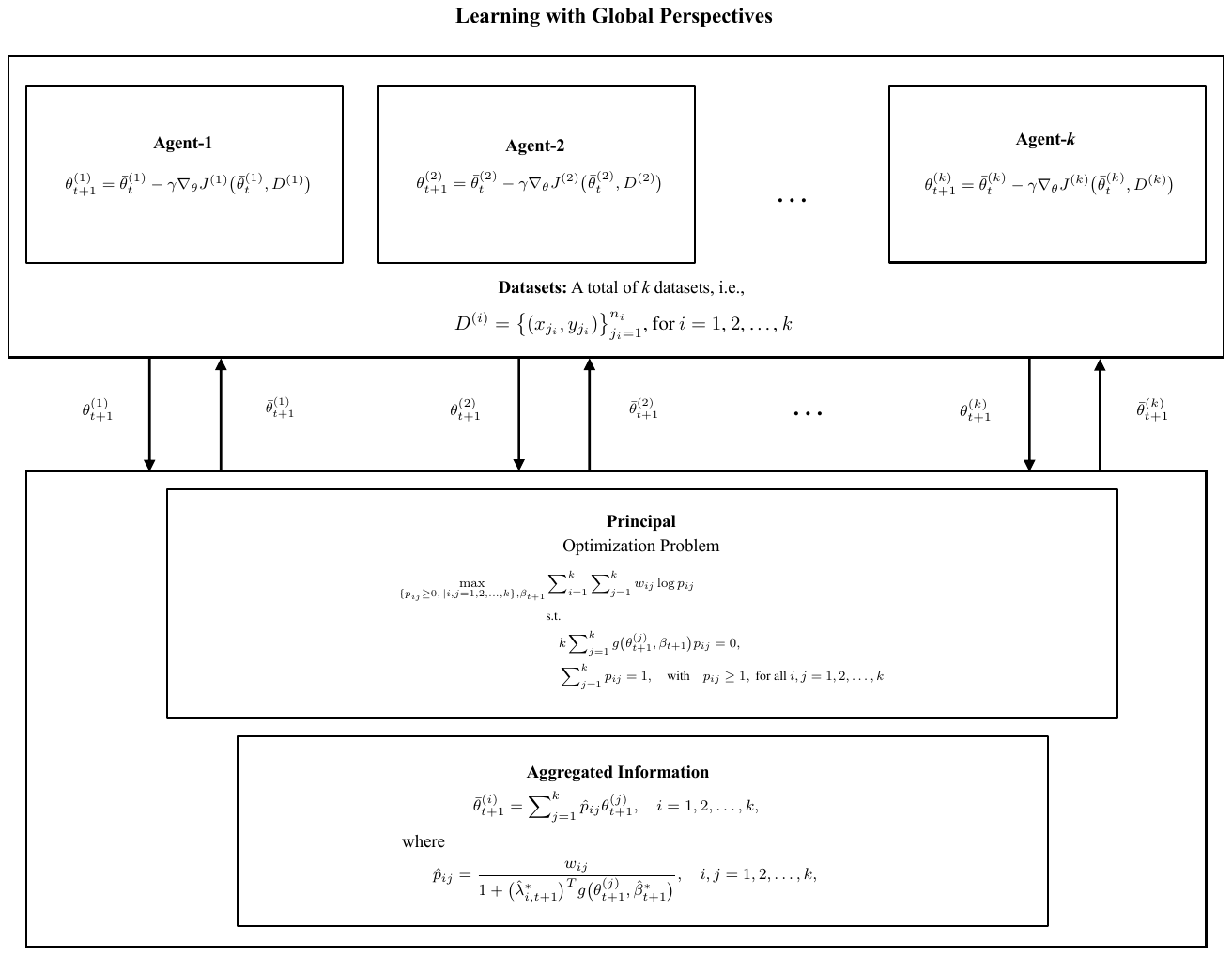}
 \caption{The functional structure for the proposed abstract learning framework.} \label{Figure1}
\end{center}
\end{figure}

In this section, we provide a formal problem statement that allows us to outline our abstract learning framework with global perspectives. In particular, the proposed learning framework, which exploits the dynamic-optimizing principal-agent setting, consists of the following core ideas: (i) {\it Datasets:} A total of $k$ datasets, i.e., $D^{(i)} = \bigl\{ (x_{j_i}, y_{j_i})\bigr\}_{j_i=1}^{n_i}$, for $i=1$, $2$, \ldots, $k$, each with a dataset size of $n_i$, with $\sum\nolimits_{i=1}^k n_i = n$, that may be generated from a given original dataset by means of bootstrapping with/without replacement, for an instance of {\it divide-and-conquer} paradigm dealing with large dataset, or they may simply represent distributed/decentralized observations associated with point estimation problem in modeling of high-dimensional nonlinear functions. (ii) {\it Agents:} There are $k$ agents, numbered as ${\rm Agent}$-1, ${\rm Agent}$-2, \ldots, ${\rm Agent}$-$k$, and they are primarily tasked to search for a set of parameter estimates $\theta \in \Theta \subseteq \mathbb{R}^{d_{\theta}}$, i.e., from a finite-dimensional parameter space, such that the function $h\bigl(x;\theta\bigr) \in \mathcal{H}$, i.e., from a given class of hypothesis function space $\mathcal{H}$, describes best the corresponding training datasets $D^{(i)}$, for $i=1$, $2$, \ldots, $k$. Here, we also assume that each agent updates its parameter estimates, with some global perspectives, using a discrete-time approximation of a gradient dynamical system, but guided by its corresponding training dataset. (iii) {\it Principal:} The principal, which is exerting an influence on the learning process of the agents in the aggregation, is tasked to solve a high-level optimization problem posed as an empirical-likelihood estimator under conditional moment restrictions model that accounts information about the agents' predictive performances on out-of-samples as well as a set of private datasets available only to the principal. Here, we also assume that the principal uses a separate data $\bigl(x_{n+1}, y_{n+1} \bigr)$ at each instant of iteration for assessing the agents' predictive performances on out-of-samples. While, the conditional moment restrictions is performed with respect to a set of private datasets $\bigl\{s_1, s_2, \ldots, s_k \bigr\}$, with support $\mathbb{R}^{d_s}$, that is available only to the principal, but provides global perspectives to the agents about the learning process, i.e., the ability to view things according to their relative importances or in their true relations. Figure~\ref{Figure1} shows a functional structure for the proposed abstract learning framework with global perspectives, where various information -- such as set of local datasets available only to the agents as part of their learning process or tasks, a separate dataset for accessing the agents' predictive performances on out-of-samples, and a set of private datasets available only to the principal for exerting influence on the learning process -- are utilized in the dynamic-optimizing principal-agent setting.

In terms of mathematical construct, searching for a set optimal parameter estimates $\theta^{\ast} \in \Theta$ can be obtained by the method of aggregating a set of {\it steady-state solutions} corresponding to the following family of gradient dynamical systems, whose {\it time-evolutions} are guided by the corresponding training datasets $D^{(i)}$, $i=1$, $2$, \ldots, $k$, i.e.,
\begin{align}
 \dot{\theta}^{(i)}(t) = - \nabla_{\theta} J^{(i)}\bigl(\theta^{(i)}(t),D^{(i)}\bigr), \quad \theta^{(i)}(0) = \theta_0, \quad \biggl(\text{with} ~~~ \dot{\theta}(t) \triangleq \frac{d}{dt}\theta(t) \biggr), \label{Eq2.1}
\end{align}
where the term $J^{(i)}(\theta^{(i)}, D^{(i)}) = \frac{1}{n_i} \sum\nolimits_{{j_i}=1}^{n_i} {\ell}\bigl(y_{j_i}, h\bigl(x_{j_i};\theta^{(i)}\bigr)\bigr)$, with a suitable loss function $\ell$ that quantifies the lack-of-fit between the assumed nonlinear functional model $h(\,\cdot\,;\theta) \in \mathcal{H}$ under investigation and the datasets $D^{(i)}$, for $i=1$, $2$, \ldots, $k$. Note that such an approach has a series short coming due to the solutions may become trapped at a local minimum, rather than the global minimum solution (e.g., see \cite{r10}). One way of addressing such difficulty is to embed the above family of gradient dynamical systems in a general abstract learning framework that fosters learning process with global perspectives as well as aggregation with an inherent feedbacks behavior and collaborations among the agents, which is the focus of this brief note.

{\bf General assumptions:} {\it Throughout this note, we assume that $\nabla_{\theta} J^{(i)}(\theta,D^{(i)})$, for each $i \in \{1,2, \ldots, k\}$, is uniformly Lipschitz and further satisfies the following growth condition 
\begin{align}
\bigl\vert \nabla_{\theta} J^{(i)}(\theta,D^{(i)}) \bigr\vert^2 \le L_{\rm Lip} \bigl(1 + \vert \theta \vert^2 \bigr), \quad \forall\, \theta \in \Theta, \label{Eq2.2}
\end{align}
for some constant $L_{\rm Lip} > 0$.}

\section{Main results} \label{S3}
In this section, we present our main results, where we present a coherent mathematical argument which is necessary for characterizing the learning process behind this abstract learning framework with dynamic-optimizing principal-agent setting.

\subsection{Parameter-updating model for the agents}\label{S3.1}
In what follows, we allow each agent to update its parameter estimate using a discrete-time approximation of the gradient system in Equation~\eqref{Eq2.1}, whose time-evolution is guided by its respective training dataset $D^{(i)}$, with $i \in \{1,2, \ldots, k\}$,, i.e.,
\begin{align}
 \theta_{t+1}^{(i)} = \bar{\theta}_{t}^{(i)} - \gamma_{t}^{(i)} \nabla_{\theta} J^{(i)}\bigl(\bar{\theta}_{t}^{(i)},D^{(i)}\bigr), \quad \text{with} \quad \bar{\theta}_{t}^{(i)} = \sum\nolimits_{j=1}^k p_{ij} \theta_{t}^{(j)}, \label{Eq3.1}
 \end{align}
where  
\begin{itemize}
\item[--\,] $\sum\nolimits_{j=1}^k p_{ij}=1$, with $p_{ij} \ge 0$, for $i,j \in \bigl\{1,\, 2,\, \ldots, k\bigr\}$ are the probability mass information (used  for aggregating the agents' parameter estimates) and shared with the agents at each iteration instant by the principal,
\item[--\,] $\bar{\theta}_{t}^{(i)} = \sum\nolimits_{j=1}^k \theta_{t}^{(j)}$, for $i = 1$, $2$, \ldots, $k$, are aggregated information used by the agents in their parameter updating model, and 
\item[--\,] $\gamma_{t}^{(i)}$ is a discretization time constant, i.e., $\gamma_{t}^{(i)} = \tau_{t+1} - \tau_t$, for $t = 0$, $1$, $2$, \ldots, $N-1$ and $i=1$, $2$, \ldots, $k$, with $0=\tau_0 < \tau_1< \ldots < \tau_t<\ldots<\tau_N=T$, of the time interval $[0,T]$.
\end{itemize}
Note that the probability mass information, with support on the parameter space as well as on a set of private datasets to the principal, are solutions arising from the high-level optimization problem which is tasked to the principal.

\subsection{Optimization problem for the principal}\label{S3.2}
In this subsection, we assume that the principal and the agents follow a {\it two-sided information sharing protocol}, i.e., a procedure on how they share their respective information between them. Suppose that, at the end of each iteration instant, all agents are simultaneously reporting their current updated parameter estimates to the principal. Then, the principal solves a high-level optimization problem which is posed as an empirical-likelihood estimator under conditional moment restrictions model that also accounts information about the agents' predictive performances on out-of-samples as well as a set of private information available only to the principal. To be more specific, suppose we have $k$ a set of independent parameter estimates $\bigl(\theta^{(1)},\,\theta^{(2)}, \ldots, \theta^{(k)})$, with support $\Theta \subseteq \mathbb{R}^{d_{\theta}}$, and we then proceed with determining/estimating a set of unknown parameters $\beta =(\mu, \sigma^2) \in \mathbb{B} \subseteq \mathbb{R} \times \mathbb{R}_{+}$ under conditional moment restrictions model
\begin{equation}
E\left[g\bigl(\bigl(x_{n+1},y_{n+1}\bigr), \theta, \beta\bigr) \,\bigl\vert\, s\right] = \left[\begin{matrix}
                     E\left[\bigl(y_{n+1} - h\bigl(x_{n+1};\theta\bigr) - \mu\bigr) \,\bigl\vert\, s\right]\\
                     E\left[ \bigl(\bigl(y_{n+1} - h\bigl(x_{n+1};\theta\bigr) - \mu\bigr)^2 - \sigma^2\bigr) \,\bigl\vert\, s\right] \end{matrix} \right] = 0, \label{Eq3.2}
\end{equation}
where the measurable function $g$ is given by
\begin{equation}
g\bigl(\bigl(x_{n+1},y_{n+1}\bigr), \theta, \beta\bigr) = \left[\begin{matrix}
                     \bigl(y_{n+1} - h\bigl(x_{n+1};\theta\bigr) - \mu\bigr)\\
                     \bigl(\bigl(y_{n+1} - h\bigl(x_{n+1};\theta\bigr) - \mu\bigr)^2 - \sigma^2\bigr) \end{matrix} \right] \in \mathbb{R}^{d_g}, \notag
 \end{equation}
while the quantity $s$, with support from $\mathbb{R}^{d_s}$, is assumed to contain information that provides  global perspectives to the agents. Note that the measurable function $g \colon (x,y) \times \Theta \times \mathbb{B} \to \mathbb{R}^{d_g}$ (with $\mathbb{R}^{d_g} \equiv \mathbb{R}^2$) provides information about prediction errors with respect to the agent's parameter estimate $\theta$.
  
In what follows, for each iteration instant $t=0$, $1$, \ldots, $N-1$, let $p_{ij}$, for each $i,j \in \{1, 2, \ldots, k\}$, be a probability mass placed at $\bigl(\theta_{t+1}^{(i)}, s_j \bigr)$ by a discrete probability distribution that has support on $\bigl\{\theta_{t+1}^{(1)},\,\theta_{t+1}^{(2)}, \ldots, \theta_{t+1}^{(k)}\bigr\} \times \bigl\{s_1,\,s_2, \ldots, s_k\bigr\}$. Further, let $w_{ij}$ be the weights given by
\begin{align}
w_{ij} = \frac{\psi\bigl(\tfrac{s_i - s_j}{h_{bw}}\bigr)}{\sum\nolimits_{j=1}^k \psi\bigl(\tfrac{s_i - s_j}{h_{bw}}\bigr)}, \quad i,j =1, 2, \ldots, k, \label{Eq3.3}
\end{align}
for some appropriately chosen kernel function $\psi$ and bandwidth $h_{bw}$. Moreover, if we define the following smoothed log-likelihood function $\ell_{\theta}\bigl(\beta_{t+1}\bigr)$,
\begin{align}
\ell_{\theta}\bigl(\beta_{t+1}\bigr) = \sum\nolimits_{i=1}^k \sum\nolimits_{j=1}^k w_{ij} \log p_{ij}. \label{Eq3.4}
\end{align}
Then, we have the following optimization problem
\begin{align}
 \max_{\{ p_{ij}\ge 0,\,\vert  i,j=1, 2, \ldots, k\}, \beta_{t+1}} & \sum\nolimits_{i=1}^k \sum\nolimits_{j=1}^k w_{ij}\log p_{ij} \label{Eq3.5}\\
                                         &\text{s.t.} \notag \\
                                         & \quad k \sum\nolimits_{j=1}^k g\bigl(\theta_{t+1}^{(j)}, \beta_{t+1}\bigr) p_{ij} = 0, \notag\\
                                         & \quad \sum\nolimits_{j=1}^k p_{ij} = 1, \quad \text{with} \quad p_{ij} \ge 1, ~\text{for all} ~ i,j = 1, 2, \ldots, k, \notag
\end{align}
that needs to be solved by the principal for each iteration instant $t=0$, $1$, \ldots, $N-1$. Note that the above optimization problem in Equation~\ref{Eq3.5} can be conveniently solved using Lagrange multipliers (i.e., based on the KKT's optimality conditions (e.g., see \cite{r11})) as follows
\begin{align}
\mathscr{L}\bigl(\beta_{t+1} \bigr) = \sum\nolimits_{i=1}^k \sum\nolimits_{j=1}^k  w_{ij}\log p_{ij} &- \sum\nolimits_{i=1}^k \eta_i \left( \sum\nolimits_{j=1}^k p_{ij} - 1 \right) \notag \\
& - k \sum\nolimits_{i=1}^k \lambda_i^T\sum\nolimits_{j=1}^k g\bigl(\theta_{t+1}^{(j)}, \beta_{t+1}\bigr) p_{ij}, \label{Eq3.6}
\end{align}
where $\lambda_i \in \mathbb{R}^{d_g}$ and $\eta_i \in \mathbb{R}$, for $i=1$, $2$, \ldots, $k$, are the Lagrange multipliers corresponding to the constraints in Equation~\ref{Eq3.5}. Then, it is easy to see that the solutions are given by
\begin{align}
\hat{p}_{ij} = \frac{w_{ij}}{1 + \lambda_i^T g\bigl(\theta_{t+1}^{(j)}, \beta_{t+1}\bigr)}, \quad i,j = 1, 2, \ldots, k, \label{Eq3.7}
\end{align}
while for each $\beta \in \mathbb{B}$, the Lagrange multiplier $\lambda_i$, for $i = 1$, $2$, \ldots, $k$, solves 
\begin{align}
\sum\nolimits_{j=1}^k \frac{w_{ij} g\bigl(\theta_{t+1}^{(j)}, \beta_{t+1}\bigr)}{1 + \lambda_i^T g\bigl(\theta_{t+1}^{(j)}, \beta_{t+1}\bigr)} = 0. \label{Eq3.8}
\end{align}
Note that, for an initial value $\hat{\beta}^0$, starting with $s=0$, and for each iteration instant $t=0$, $1$, \ldots, $N-1$, the following generic algorithm, that iterates between the following two coupled optimization steps
\begin{align}
\hat{\lambda}_i^{s+1} = \argmax_{\lambda \in \mathbb{R}^{d_g}} \sum\nolimits_{j=1}^k w_{ij} \log \bigl(1 + \lambda_i^T g\bigl(\theta_{t+1}^{(j)}, \hat{\beta}^s\bigr), \quad i = 1, 2, \ldots, k \label{Eq3.9}
\end{align}
and
\begin{align}
\hat{\beta}^{s+1} = \argmin_{\beta \in \mathbb{B}} - \sum\nolimits_{i=1}^k \sum\nolimits_{j=1}^k w_{ij} \log \bigl(1 + \bigl(\hat{\lambda}_i^{s+1}\bigr)^T g\bigl(\theta_{t+1}^{(j)}, \beta\bigr)\bigr) \label{Eq3.10}
\end{align}
can be used for determining a set of optimal solutions $\hat{\lambda}_{i,t+1}^{\ast}$, for $i=1$, $2$, \ldots, $k$, and $\hat{\beta}_{t+1}^{\ast}$. Note that, during the next iteration cycle, based on $\bigl(\bigl\{\hat{\lambda}_{i,t+1}^{\ast}\bigr\}_{i=1}^k, \hat{\beta}_{t+1}^{\ast} \bigr)$, $\bigl\{\theta_{t+1}^{(i)}\bigr\}_{i=1}^k$ and $\bigl\{w_{i,j}\bigr\}_{i,j=1}^k$, the principal then shares information with each of the agents in the form of some aggregated information, i.e.,
\begin{align}
\bar{\theta}_{t+1}^{(i)} = \sum\nolimits_{j=1}^k \hat{p}_{ij} \theta_{t+1}^{(j)}, \quad i = 1, 2, \ldots, k, \label{Eq3.11}
\end{align}
with
\begin{align*}
\hat{p}_{ij} = \frac{w_{ij}}{1 + \bigl(\hat{\lambda}_{i,t+1}^{\ast}\bigr)^T g\bigl(\theta_{t+1}^{(j)}, \hat{\beta}_{t+1}^{\ast}\bigr)}, \quad i,j = 1, 2, \ldots, k.
\end{align*}
Finally, it is worth remarking that, when $t = N-1$, the log-likelihood function $\ell_{\theta}\bigl(\hat{\beta}_{N}^{\ast}\bigr)$ satisfies the following upper bound
\begin{align}
\ell_{\theta}\bigl(\beta_{N}^{\ast}\bigr) &= \sum\nolimits_{i=1}^k \sum\nolimits_{j=1}^k w_{ij} \log \hat{p}_{ij} \notag\\
                                                  &= \sum\nolimits_{i=1}^k \sum\nolimits_{j=1}^k \log \left(\frac{w_{ij}}{1 + \bigl(\hat{\lambda}_{i,N}^{\ast}\bigr)^T g\bigl(\theta_{N}^{(j)}, \hat{\beta}_{N}^{\ast}\bigr)}\right) \notag\\
                                                  &= \sum\nolimits_{i=1}^k \sum\nolimits_{j=1}^k \log w_{ij} - \sum\nolimits_{i=1}^k \sum\nolimits_{j=1}^k \log\left(1 + \bigl(\hat{\lambda}_{i,N}^{\ast}\bigr)^T g\bigl(\theta_{N}^{(j)}, \beta_{N}^{\ast}\bigr)\right) \notag\\
                                                   &\le - \sum\nolimits_{i=1}^k \sum\nolimits_{j=1}^k \log\left(1 + \bigl(\hat{\lambda}_{i,N}^{\ast}\bigr)^T g\bigl(\theta_{N}^{(j)}, \beta_{N}^{\ast}\bigr)\right), \label{Eq3.12}
\end{align}
since $\sum\nolimits_{i=1}^k \sum\nolimits_{j=1}^k \log w_{ij} \le 0$, due to $w_{ij} \in (0, 1]$.

\subsection{Algorithm} \label{S3.3} In this subsection, we present a generic algorithm that provides a viable way for implementing such a learning problem with global perspectives.

{\rm \footnotesize

{\bf ALGORITHM:} Learning problem with global  perspectives.
\begin{itemize}
\item[] {\bf Inputs:} {\bf (i) Datasets for the agents:} $k$ datasets, each with data size $n_i$, $D^{(i)} = \bigl\{ (x_{j_i}, y_{j_i})\bigr\}_{j_i=1}^{n_i}$, with $i=1,2, \ldots, k$; {\bf (ii) An equidistant discretization time:} $\gamma=\tau_{t+1} - \tau_t = T/N$, for $t=0,1,2, \ldots, N-1$, with $0=\tau_0 < \tau_1< \ldots < \tau_t<\ldots<\tau_N=T$, of the time interval $[0,T]$; and {\bf (iii) Datasets for the principal:} A set of private datasets  $\bigl\{s_i\bigr\}_{i=1}^{k}$ and an additional data $\bigl(x_{n+1}, y_{n+1} \bigr)$.\\

\item[{\bf 0.}] {\bf Initialize:} Set $t=0$, $\theta_0^{(i)} =\theta_0$ and $\hat{p}_{ij} = 1/k$, for $i=1$, $2$,\ldots, $k$. Compute the weights using $w_{ij} = \psi\bigl(\tfrac{s_i - s_j}{h_{bw}}\bigr)\biggl/\sum\nolimits_{j=1}^k \psi\bigl(\tfrac{s_i - s_j}{h_{bw}}\bigr)$, for $i,j= 1$, $2$, \ldots, $k$, with an appropriate kernel function and bandwidth $h_{bw}$.
\item[{\bf 1.}] {\bf Agents:} Update parameter estimates using
\begin{align*}
 \theta_{t+1}^{(i)} = \bar{\theta}_{t}^{(i)} - \gamma \nabla_{\theta} J^{(i)}\bigl(\bar{\theta}_{t}^{(i)},D^{(i)}\bigr), \quad \text{with} \quad \bar{\theta}_{t}^{(i)} = \sum\nolimits_{j=1}^k \hat{p}_{ij} \theta_{t}^{(j)},
 \end{align*}
for each $i \in \{1,2, \ldots, k\}$.
\item[{\bf 2.}] {\bf Principal:} Solves the following optimization problem
\begin{align*}
 \max_{\{ p_{ij}\ge 0,\,\vert  i,j=1, 2, \ldots, k\}, \beta_{t+1}} & \sum\nolimits_{i=1}^k \sum\nolimits_{j=1}^k w_{ij}\log p_{ij}\\
                                         &\text{s.t.} \notag \\
                                         & \quad k \sum\nolimits_{j=1}^k g\bigl(\theta_{t+1}^{(j)}, \beta_{t+1}\bigr) p_{ij} = 0,\\
                                         & \quad \sum\nolimits_{j=1}^k p_{ij} = 1, \quad \text{with} \quad p_{ij} \ge 1, ~\forall ~ i,j \in \bigl\{1, 2, \ldots, k\bigr\}
\end{align*}
by iterating between Step~$2$\,(i) and Step~$2$\,(ii), with an initial value $\hat{\beta}^0$, starting from $s=0$, involving the following coupled optimization problems:
\begin{itemize}
\item[{\bf 2\,(i).}] First compute $\bigl\{\hat{\lambda}_{i,t+1}^{\ast}\bigr\}_{i=1}^k$ using
\begin{align*}
\hat{\lambda}_i^{s+1} = \argmax_{\lambda \in \mathbb{R}^{d_g}} \sum\nolimits_{j=1}^k w_{ij} \log \bigl(1 + \lambda_i^T g\bigl(\theta_{t+1}^{(j)}, \hat{\beta}^s\bigr), \quad i = 1, 2, \ldots, k.
\end{align*}
\item[{\bf 2\,(ii).}] Then compute $\hat{\beta}_{t+1}^{\ast}$ using
\begin{align*}
\hat{\beta}^{s+1} = \argmin_{\beta \in \mathbb{B}} - \sum\nolimits_{i=1}^k \sum\nolimits_{j=1}^k w_{ij} \log \bigl(1 + \bigl(\hat{\lambda}_i^{s+1}\bigr)^T g\bigl(\theta_{t+1}^{(j)}, \beta\bigr)\bigr).
\end{align*}
\item[{\bf 2\,(iii).}] Repeat Step~$2$\,(i) and Step~$2$\,(ii) until convergence, i.e., $\left( \bigl \{\hat{\lambda}_{i,t+1}^s\bigr\}_{i=1}^k, \hat{\beta}_{t+1}^s \right)$ converges to $\left(\bigl\{\hat{\lambda}_{i,t+1}^{\ast}\bigr\}_{i=1}^k, \hat{\beta}_{t+1}^{\ast} \right)$,
\end{itemize}
\item[{\bf 3.}] Then, compute the aggregated information, i.e.,
\begin{align*}
\bar{\theta}_{t+1}^{(i)} = \sum\nolimits_{j=1}^k \hat{p}_{ij} \theta_{t+1}^{(j)}, \quad i = 1, 2, \ldots, k,
\end{align*}
with
\begin{align*}
\hat{p}_{ij} = \frac{w_{ij}}{1 + \bigl(\hat{\lambda}_{i,t+1}^{\ast}\bigr)^T g\bigl(\theta_{t+1}^{(j)}, \hat{\beta}_{t+1}^{\ast}\bigr)}, \quad i,j = 1, 2, \ldots, k,
\end{align*}
and further share with each of the agents for the next round iteration.
\item[{\bf 4.}] Increment $t$ by $1$ and, then repeat Steps $1$, $2$ and $3$ until convergence, i.e., $\Vert \ell_{\theta}\bigl(\hat{\beta}_{t+1}^{\ast} \bigr) - \ell_{\theta}\bigl(\hat{\beta}_{t}^{\ast} \bigr)\Vert \le {\rm tol}$, or $t=N-1$.
 \item[] {\bf Outputs:} {\bf (i)} Optimal parameter estimates: $\bar{\theta}_{N}^{(i)} = \sum\nolimits_{j=1}^k \hat{p}_{ij} \theta_{N}^{(j)}$, for $i = 1$, $2$, \ldots, $k$, and {\bf (ii)} The maximum log-likelihood: $\ell_{\theta}\bigl(\hat{\beta}_{N}^{\ast}\bigr)= \sum\nolimits_{i=1}^k \sum\nolimits_{j=1}^k w_{ij} \log \hat{p}_{ij} $.
\end{itemize}}

\end{document}